\documentclass[wcp,x11names,table]{jmlr}

\usepackage{booktabs}

\usepackage[plain]{fancyref}
\usepackage{algorithm}
\usepackage[all]{xy}
\usepackage{algorithmic}
\usepackage{graphicx}
\usepackage{color}         
\definecolor{webgreen}{rgb}{0,.5,0}

\newcommand*{\fancyrefthmlabelprefix}{alg}
\frefformat{plain}{\fancyrefthmlabelprefix}{algorithm~#1}
\Frefformat{plain}{\fancyrefthmlabelprefix}{Algorithm~#1}

\jmlrvolume{0}
\jmlryear{0}
\jmlrworkshop{0}

\title[Temporal Autoencoding]{Temporal Autoencoding Improves Generative Models of Time Series}

  \author{\Name{Chris H\"ausler\footnotemark[1], Martin P. Nawrot}\\ \Email{chausler@gmail.com, martin.nawrot@fu-berlin.de}\\ \addr{Neuroinformatics, Freie Universit\"at Berlin} \AND \Name{Alex Susemihl\footnotemark[1], Manfred Opper}\\ \Email{alexsusemihl@gmail.com, opperm@cs.tu-berlin.de}\\ \addr{Artificial Intelligence, Technische Universit\"at Berlin}
  \\
  \\
 \footnotemark[1] These authors have contributed equally to this work.
 }

\begin{document}

\maketitle

\begin{abstract} 
Restricted Boltzmann Machines (RBMs) are generative models which can learn useful representations from samples of a dataset in an unsupervised fashion. They have been widely employed as an unsupervised pre-training method in machine learning. RBMs have been modified to model time series in two main ways: The Temporal RBM stacks a number of RBMs laterally and introduces temporal dependencies between the hidden layer units; The Conditional RBM, on the other hand, considers past samples of the dataset as a conditional bias and learns a representation which takes these into account. Here we propose a new training method for both the TRBM and the CRBM, which enforces the dynamic structure of temporal datasets. We do so by treating the temporal models as denoising autoencoders, considering past frames of the dataset as corrupted versions of the present frame and minimizing the reconstruction error of the present data by the model. We call this approach Temporal Autoencoding. This leads to a significant improvement in the performance of both models in a filling-in-frames task across a number of datasets. The error reduction for motion capture data is 56\% for the CRBM and 80\% for the TRBM. Taking the posterior mean prediction instead of single samples further improves the model's estimates, decreasing the error by as much as 91\% for the CRBM on motion capture data. We also trained the model to perform forecasting on a large number of datasets and have found TA pretraining to consistently improve the performance of the forecasts.
Furthermore, by looking at the prediction error across time, we can see that this improvement reflects a better representation of the dynamics of the data as opposed to a bias towards reconstructing the observed data on a short time scale.
We believe this novel approach of mixing contrastive divergence and autoencoder training yields better models of temporal data, bridging the way towards more robust generative models of time series.

\end{abstract}

\section{Introduction}
The statistical modelling of temporal data is a problem of great interest to machine learning. Not only because of the number of data sources that are intrinsically temporal, but also because of the growing number of applications that interact with users in real time and require efficient and scalable handling of large streams of temporal data. Good models of the statistical structure of datasets are also generally thought to yield good representations for discriminative or predictive tasks on these data. One class of statistical model which has received a great deal of attention in the recent literature is the Restricted Boltzmann Machine \citep{hinton2006reducing}. The Restricted Boltzmann Machine (RBM) is a simple graphical model which is easily trainable using contrastive divergence (CD) learning \citep{carreira2005contrastive}.
There are two canonical ways in which RBMs have been extended to model temporal data: The Temporal RBM \citep{sutskever2007learning}; and the Conditional RBM \citep{taylor2007modeling}, both of which have had notable success. The TRBM learns temporal correlations between latent representations for each temporal sample, while the CRBM learns a latent representation for the whole data sequence (see \fref{sec:methods} below).\par

One marked advantage of these methods is that they allow for the generation of samples from the learned data distribution.
However, although these methods have had success generating data from a number of data sources, they have only done so in relatively narrow contexts. Here we improve on the training methods of temporal and conditional RBMs to allow for better and more robust generation from general datasets.\par

The TRBM and CRBM seek to model the structure of the data, but the learning methods usually employed disregard the  causality in its structure. For these models, contrastive divergence learning seeks to approximately maximize the likelihood of sequences of observed data (in the case of the TRBM) or the conditional likelihood of the present data given the past (in the case of the CRBM), without any regard to the underlying dynamics. Naturally, the method learns a dynamical model of the data, but to explicitly train it do so could result in better models of the system's dynamics. We propose a simple method to enforce the dynamics of the data in the models learnt representations. We achieve this by training the model as a neural network for prediction, similar to what is done in denoising autoencoders \citep{Vincent2010}. We refer to this approach as Temporal Autoencoding (TA), which by itself it does not yield good generative models. However, by initializing the model through Temporal Autoencoding and then applying contrastive divergence training, one can bias the structure of the models towards the dynamics of the data, resulting in better generative performance.\par
One natural way to measure the quality of a generative model is to take samples from it and compare them to samples from the dataset. A simple way to quantify the fidelity of these samples is to provide partial samples from the data where certain dimensions are left out and then to generate the missing dimensions by sampling from the model. This approach is generally called \emph{filling-in} and is particularly well-suited to temporal applications as we can condition on the observations up to a certain time and fill in the missing frames by sampling from the model. One can then compare the generated sample to the true data, for example by taking the mean squared error (MSE)or the Mean Absolute Percentage Error (MAPE) between them. We will use these measures in a \emph{filling-in-frames} task to quantify thhe quality of our models throughout this paper.\par
Temporal Autoencoding pretraining improves the performance of both generative models across all datasets considered by as much as 80\% with approximately the same training time those models trained in the conventional manner. These findings hold across different modalities of data, such as human motion capture data, and a number of datasets taken from the M3 forecasting competition \citep{makridakis2000m3}, which encompass yearly, quarterly, monthly and a few unspecified types of temporal data . The fact that the proposed pretraining betters model performance across datasets for both the CRBM and TRBM confirms that the method provides a robust improvement in the generative performance of both RBM models. Furthermore, the performance increase is not limited to short time-scales, but can be seen to hold even for longer periods of time ranging over the memory encoded directly by the method.\par
Autoencoders have recently been cast into a new light by considering them as generative models \citep{Bengio2013generative}. Though we do not take that approach here, we firmly believe that autoencoder training can improve the performance of generative models greatly. This has been shown for the temporal models considered here, and we expect this to lead to a significant improvement towards training temporal generative models.

\section{Methods}

\label{sec:methods}

We propose a new pretraining method for both the TRBM and the CRBM, based on a denoising autoencoder approach through time. To this end we shortly discuss the RBM, the denoising autoencoder and the temporal models used. Throughout the paper we will denote the activation of visible layers by $\mathbf{v} = (v_1,v_2,\ldots,v_N)$ and the activation of hidden layers by $\mathbf{h} = (h_1,h_2,\ldots,h_M)$, where $N$ is the number of visible units and $M$ the number of hidden units. In the case of temporal models we will denote the present state of the visible and hidden layers by $\mathbf{v}^T= (v^T_1,v^T_2,\ldots,v^T_N)$ and $\mathbf{h}^T = (h^T_1,h^T_2,\ldots,h^T_M)$, where $T$ is the number of delayed units considered, and the subsequential delayed units by $\mathbf{v}^k = (v^k_1,v^k_2,\ldots,v^k_N)$ and $\mathbf{h}^k = (h^k_1,h^k_2,\ldots,h^k_M)$, where $k \in \{0,\ldots,T-1\}$. The naming convention is shown in \fref{fig:all_models} for $T=2$ delayed units. 

\subsection{Restricted Boltzmann Machines}

Restricted Boltzmann Machines are generative models which assume all-to-all symmetric connectivity between the visible and hidden variables (see \fref{fig:all_models}a) and seek to model the structure of a given dataset. They are energy-based models, parametrized by a $N$-by-$M$-dimensional weight matrix $\mathbf{W}$, a bias for the visible layer $\mathbf{b}^v = (b^v_1,b^v_2,\ldots,b^v_N)$ and a bias for the hidden layer $\mathbf{b}^h =(b^h_1,b^h_2,\ldots,b^h_M)$. The energy of a given configuration of activations $\mathbf{v}$ and $\mathbf{h}$ is given by
$$
E_{RBM}(\mathbf{v},\mathbf{h}|\mathbf{W},\mathbf{b}^v,\mathbf{b}^h) = -\sum_{i,j} W_{ij} \,v_i\, h_j - \sum_i b^v_i\, v_i -\sum_j b^h_j \,h_j,
$$
and the probability of a given configuration is given by
$$
P(\mathbf{v},\mathbf{h}) = \exp\left(-E_{RBM}(\mathbf{v},\mathbf{h}|\mathbf{W},\mathbf{b}^v,\mathbf{b}^h)\right)/Z(\mathbf{W,b}_v,\mathbf{b}^h),
$$
where $Z(\mathbf{W,b}_v,\mathbf{b}^h)$ is the partition function. One noted advantage of the RBM is that the visible units are independent of each other when conditioned on the hidden units and vice-versa. This allows for efficient sampling, and for the exact calculation of a number of averages. Namely, we can evaluate exactly the conditional distributions
\[
P(v_i = 1| \mathbf{h}) = \sigma\left(\sum_{j} W_{ij} h_j + b^v_i\right),
\]
and
\[
P(h_j = 1|\mathbf{v}) = \sigma\left(\sum_{i} W_{ij} v_i + b^h_j\right),
\]
where $\sigma(x ) = 1/(1+\exp(-x))$ is the sigmoid function.\par
 One can extend the RBM to continuous-valued visible variables by modifying the energy function, to obtain the Gaussian-binary RBM
$$
E_{RBM}(\mathbf{v},\mathbf{h}|\mathbf{W},\mathbf{b}^v,\mathbf{b}^h,\{\sigma_i^2\}
) = - \sum_{i,j} \frac{1}{\sigma_i^2} W_{ij}\, v_i \,h_j
 +\sum_i \frac{(b^v_i- v_i)^2}{2\sigma_i^2}  -\sum_j b^h_j \,h_j.
$$
This then leads to the conditional distributions
\[
P(v_i| \mathbf{h}) = \mathcal{N}\left(\sum_{j} W_{ij} h_j + b^v_i\,,\,\sigma_i^2\right),
\]
where $\mathcal{N}(\mu,\sigma^2)$ is the normal distribution with mean $\mu$ and variance $\sigma^2$
and
\[
P(h_j = 1|\mathbf{v}) = \sigma\left(\sum_{i} \frac{W_{ij} v_i}{\sigma_i^2} + b^h_j\right).
\]
Often the variances are constrained to have the same value across dimensions, or simply taken to be constant. To learn them from the data, however, one must take extra care to deal with vanishingly small variances.
Like most statistical models, RBMs can be trained by maximizing the log likelihood of the data. This, however proves to be intractable even for the case of the RBM, and we are left with maximizing surrogate functions. The derivative of the log likelihood of an observed visible state $D$ can be written as
\[
\frac{\partial \log P(D)}{\partial \theta} = -\left<\frac{\partial E}{\partial \theta}\middle| D\right>_\mathbf{h} +  \left<\frac{\partial E}{\partial \theta}\right>_{\mathbf{v,h}},
\]
where $\theta$ is any of the parameters of the model.
Note that the first term is easy to compute, but the second one involves averages over the full distribution $P(\mathbf{v,h})$, which is intractable.
RBMs are therefore usually trained through contrastive divergence, which approximately follows the gradient of the cost function
\begin{eqnarray*}
CD_n (\mathbf{W},\mathbf{b}^v,\mathbf{b}^h) =& KL(P_0(\mathbf{v}| \mathbf{W},\mathbf{b}^v,\mathbf{b}^h)||P( \mathbf{v}|\mathbf{W},\mathbf{b}^v,\mathbf{b}^h))\\
&-KL(P_n(\mathbf{v}| \mathbf{W},\mathbf{b}^v,\mathbf{b}^h)||P( \mathbf{v}|\mathbf{W},\mathbf{b}^v,\mathbf{b}^h)),
\end{eqnarray*}
where $P_0$ is the data distribution, $P_n$ is the distribution of the visible layer after $n$ Markov chain Monte Carlo (MCMC) steps and $KL()$
 is the Kullback-Leibler divergence \citep{carreira2005contrastive}. The samples from the data distribution are simply taken from the data, whereas the samples from $P_n$ are taken by running a MCMC for $n$ steps. The function $CD_n$ gives an approximation to maximum-likelihood (ML) estimation of the weight matrix $\mathbf{W}$. Further approximation is still needed, as the $CD_n$ cost still involves intractable averages, but it is generally found that the approximate parameter update given by
$$
\Delta \theta \propto -\left<\frac{\partial E_{RBM}}{\partial \theta}\middle|D\right>_{\mathbf{h}} + \left<\frac{\partial E_{RBM}}{\partial \theta}\right>_{n},
$$
already gives very good results.
The weight updates then become
$$
\Delta W_{ij} \propto \frac{1}{\sigma_i^2}\left<v_i h_j\right>_{0}-\frac{1}{\sigma_i^2}\left<v_i h_j\right>_{n}.
$$
{In general, $n=1$ is already sufficient for practical purposes \citep{hinton2006reducing}.
\par
\subsection{Autoencoders} 
Autoencoders are deterministic models with two weight matrices $\mathbf{W}^{1}$ and $\mathbf{W}^{2}$ representing the flow of data from the visible-to-hidden and hidden-to-visible layers respectively (see Figure \ref{fig:all_models}b).\footnote{Often one only uses one matrix and propagates up throught $\mathbf{W}^1$ and down through its transpose $(\mathbf{W}^1)^\top$} AEs are trained to perform optimal reconstruction of the visible layer, often by minimizing the mean-squared error (MSE) in a reconstruction task. This is usually evaluated as follows: Given an activation pattern in the visible layer $\mathbf{v}$, we evaluate the activation of the hidden layer by $h_j = \sigma( \sum_i W^1_{ij} v_i+ b^h_j)$. These activations are then propagated back to the visible layer through $\hat{v}_i(v_i) = \sigma(\sum_j W^2_{ij} h_j + b^{v}_i)$ and the weights $\mathbf{W}^{1}$ and $\mathbf{W}^{2}$ are trained to minimize the distance measure between the original and reconstructed visible layers. Therefore, given a set of $Q$ image samples $\{D_k\}$ we can define the cost function.
For example, using the squared euclidean distance between the original data and the reconstructed data, $\hat{\mathbf{v}}_k =( \hat{v}_1(D_k), \hat{v}_2(D_k),\ldots,\hat{v}_N(D_k))$, we have the loss function
$$
\mathcal{L}(\mathbf{W}^1,\mathbf{W}^2,\mathbf{b}^v,\mathbf{b}^h| \{D_k\}) = \frac{1}{Q}\sum_{d} \| D_k - \hat{\mathbf{v}}(D_k)\|^2.
$$
The weights can then be learned through stochastic gradient descent on the cost function. Autoencoders often yield better representations when trained on corrupted versions of the original data, performing gradient descent on the distance to the uncorrupted data. This approach is called a denoising autoencoder \citep{Vincent2010}. Note that in the AE, the activations of all units are continuous and not binary, and usually take values between $0$ and $1$.
\par

\begin{figure*}
\begin{displaymath}
\entrymodifiers = {+[F-:<3pt>]}
\xymatrix{
*{\textrm{a)}}&{\mathbf{h}} &
*{\textrm{b)}}&\mathbf{h}\ar@/_/[d]_{\mathbf{W}^2}&
*{\textrm{c)}}& \mathbf{h}^0 \ar@/^1pc/[rr]^{\mathbf{W}^2} & \mathbf{h}^1\ar@/^/[r]_{\mathbf{W}^1}& \mathbf{h}^2\ar@/_1pc/[ll] 
\ar@/_/[l]&
*{\textrm{d)}}&*{}&*{}&\mathbf{h}
\\
*{}&\mathbf{v} \ar@{<->}[u]_{\mathbf{W}}&*{}
& \mathbf{v}\ar@/_/[u]_{\mathbf{W}^1} &*{}
&\mathbf{v}^0 \ar@{<->}[u]_{\mathbf{W}}&\mathbf{v}^1 \ar@{<->}[u]_{\mathbf{W}}&\mathbf{v}^2 \ar@{<->}[u]_{\mathbf{W}}
&*{}&\mathbf{v}^0\ar@/^/[urr]^{\mathbf{W}^2} \ar@/_1pc/[rr]_{\mathbf{P}^2}&\mathbf{v}^1\ar@/^/[ur]_{\mathbf{W}^1} \ar@/_/[r]^{\mathbf{P}^1} & \mathbf{v}^2\ar@{<->}[u]_{\mathbf{W}}
\\
*{}&*{}&*{}&*{}&*{}&*{}\ar@{->}[rr]_t&*{}&*{}&*{}&*{}\ar@{->}[rr]_t&*{}&*{}&*{}
}
\end{displaymath}
\caption{Described model architectures: a) RBM; b) Autoencoder; c) Temporal RBM and d) Conditional RBM.}
\label{fig:all_models} 
\end{figure*}
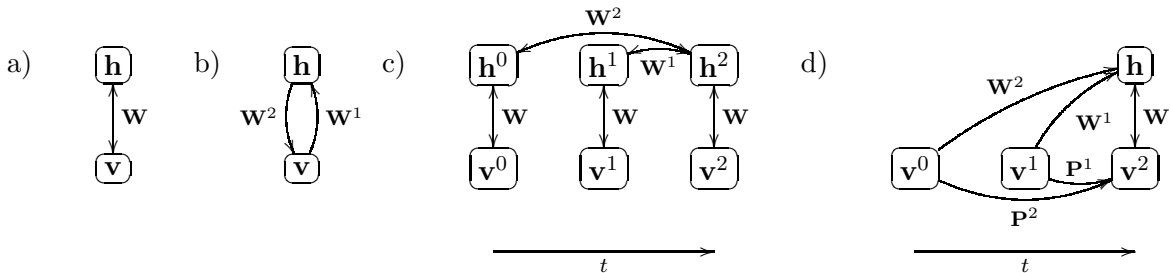

\subsection{Temporal Restricted Boltzmann Machine}
{Temporal Restricted Boltzmann Machines (TRBM)} are a temporal extension of the standard RBM whereby connections are included from previous time steps between hidden layers, from visible to hidden layers and from visible to visible layers. Learning is conducted in the same manner as a normal RBM using contrastive divergence and it has been shown that such a model can be used to learn non-linear system evolutions such as the dynamics of a ball bouncing in a box \citep{sutskever2007learning}. A more restricted version of this model, discussed in \citep{sutskever2008recurrent} can be seen in \fref{fig:all_models}c and only contains temporal connections between the hidden layers. We restrict ourselves to this model architecture throughout the paper.
\par
The energy of the model for a given configuration of the visible layers $\mathcal{V} =\{\mathbf{v}^0, \ldots , \mathbf{v}^T\}$ and hidden layers $\mathcal{H} = \{\mathbf{h}^0,\ldots, \mathbf{h}^T\}$ is given by
\begin{equation}
E(\mathcal{H,V|W,B}) = \sum_{t=0}^T E_{RBM}(\mathbf{ h}^{t},\mathbf{v}^{t}|\mathbf{W,b}^v,\mathbf{b}^h) - \sum_{t=0}^{T-1} \left(\sum_{jk} W^{(T-t)}_{jk}h^T_j  h^t_k\right),
\label{eq:energy_trbm}
\end{equation}
where we have used $\mathcal{B} = \{\mathbf{b}^v, \mathbf{b}^h\}$ and $\mathcal{W} = \{\mathbf{W},\mathbf{W}^1, \ldots, \mathbf{W}^{T}\}$, where $\mathbf{W}$ are the static weights and $\mathbf{W}^{1}, \mathbf{W}^2,\ldots \mathbf{W}^{T}$ are the delayed weights for the temporally delayed hidden layers $\mathbf{h}^{T-1},\mathbf{h}^{T-2},\ldots,\mathbf{h}^0$ (see \fref{fig:all_models}c). Note that because the hidden layers are coupled, the expectations in the CD cost can not be simply evaluated as in the RBM, and must be estimated by MCMC sampling, making training and sampling in this model more difficult.
More specifically note that the conditional distribution $P(\mathcal{H|V})$ is already intractable. A simple way to deal with this is the so-called filtering approximation, where we sample from the past hidden layers ignoring the present hidden layer and then sample from the present hidden layer conditioned on the past.

\subsection{Conditional Restricted Boltzmann Machines} One way to overcome the problems of the TRBM has been proposed in the Conditional Restricted Boltzmann Machines \citep{taylor2007modeling}. The CRBM has only one hidden layer, which receives input from all visible layers, past and present. Additionally, the present visible layer receives input from past visible layers. Unlike the TRBM, only the present hidden and visible layers are considered to be free, whereas the past visible states are conditioned on. The energy of the model can be written as
\begin{eqnarray*}
&E_{CRBM}(\mathbf{h}^T,\mathbf{v}^T|\mathbf{v}^0,\ldots,\mathbf{v}^{T-1},\mathcal{W,B,P}) = E_{RBM}(\mathbf{h}^T,\mathbf{v}^T| \mathbf{W,b}_h,\mathbf{b}^v) \\
&- \sum_{t=0}^{T-1} \left( \sum_{ij} W^{(T-t)}_{ij}\,v^t_i\, h^T_j + \sum_{il} P^{(T-t)}_{il}\,v^t_i\, v^T_l\right),
\end{eqnarray*}
where $\mathcal{P} = \{P_0,\ldots P_{T-1}\}$ are the visible-to-visible weights.
The model architecture can be seen in Figure \ref{fig:all_models}d. Using this formulation, the hidden layer can still be easily marginalized over, allowing for more efficient training using contrastive divergence. The CRBM is possibly the most successful of the temporal RBM models to date and has been shown to both model and generate data from complex dynamical systems such as human motion capture data and video textures \citep{taylor2009composable}.

\subsection{Temporal Autoencoding Training}
\label{sec:atrbm_training}

The usual CD training for the TRBM and CRBM seeks to maximize the likelihood of the data observed. This usually works quite well and has been shown to allow the trained models to reproduce complex temporal data such as video  of a bouncing ball or human motion capture. However, there is one bit of essential information which these training methods overlook. They ignore that the current time frame has a causal dependence on the previous frames. If the data comes from a time series it is a natural assumption that the future states are given by some function of the past states, latent variables and possibly noise. We seek to explore this property, by explicitly learning a representation which represents these dynamics.\par
We do so by treating the hidden layers of the model as an information bottleneck, similar to what is done in the training of the denoising autoencoder \citep{Vincent2010}. We treat the past states of the time series up to a number of delays as a noisy representation of the present state, and propagate the activations through the model, considering it as a neural network with sigmoidal activation functions and perform gradient descent on the quadratic error of the reconstructed present state. In this way, we explicitly constrain the model to represent the dynamic structure of the data.\par
This essentially amounts to performing supervised learning for reconstruction using the architectures shown in \fref{fig:autoencoding}. Though the idea behind the training procedure is the same for both models, the specifics are slightly different and as such we consider them separately below.
\subsubsection{Temporal Autoencoding for the TRBM}

Let us first consider the TRBM. The energy of the model is given by \fref{eq:energy_trbm} and is essentially an $T$-th order autoregressive RBM which is usually trained by standard contrastive divergence. Here we propose to train it with a novel approach, highlighting the temporal structure of the stimulus.  First, the individual RBM visible-to-hidden weights $\mathbf{W}$ are initialized through contrastive divergence learning with a sparsity constraint on static samples of the dataset. 
After that, to ensure that the weights representing the hidden-to-hidden connections ($\mathbf{W}^{t}$) encode the dynamic structure of the ensemble, we initialize them by pre-training in the fashion of a denoising Autoencoder.
For that, we consider the model to be a deterministic Multi-Layer Perceptron with continuous activation in the hidden layers. We then consider the $T$ delayed visible layers as features and try to predict the current visible layer by projecting through the hidden layers. In essence, we are considering the model to be a feed-forward network, where the delayed visible layers would form the input layer, the delayed hidden layers would constitute the first hidden layer, the current hidden layer would be the second hidden layer and the current visible layer would be the output as is pictured in \fref{fig:autoencoding}. Given sample activations of the visible layers given by $\mathcal{V}_d =\{\mathbf{v}^0_d,\mathbf{v}^1_d, \ldots , \mathbf{v}^{T-1}_d, \mathbf{v}^{T}_d\}$, we can then write the prediction of the network as $\mathbf{\hat{v}}^T(\mathbf{v}^0_d,\mathbf{v}^1_d, \ldots , \mathbf{v}^{T-1}_d; \mathcal{W,B})$, where the $d$ index runs over the $Q$ data points. The exact format of this function is described in \fref{alg:pretraining}. We therefore minimize the reconstruction error given by
$$
\mathcal{L}(\mathcal{W,B}) = \frac{1}{Q}\sum_d \left\| \mathbf{v}_d^T -\mathbf{\hat{v}}^T(\mathbf{v}^0_d,\mathbf{v}^1_d, \ldots , \mathbf{v}^{T-1}_d; \mathcal{W,B})\right\|^2,
$$
where the sum over $d$ goes over the entire dataset. 
 After the Temporal Autoencoding is completed, the whole model (both visible-to-hidden and hidden-to-hidden weights) is trained together using contrastive divergence (CD) training. A summary of the training method is described in \fref{tab:training}.\par

\begin{table}[th]
\caption{Autoencoded TRBM Training Steps\label{tab:training}}
\begin{center}
\begin{tabular}{l|p{8cm}}
{\bf Step} \cellcolor[gray]{0.7} &{\bf Action} \cellcolor[gray]{0.7}
\\ \hline 
1. Static RBM Training \cellcolor[gray]{0.95} & Constrain the static weights $\mathbf{W}$ using CD on single frame samples of the training data \cellcolor[gray]{0.95} \\ 
2. Temporal Autoencoding \cellcolor[gray]{0.9} & Constrain the temporal weights $\mathbf{W}^{1}$ to $\mathbf{W}^{T}$ using a denoising autoencoder on multi-frame samples of the data \cellcolor[gray]{0.9} \\   
3. Model Finalisation \cellcolor[gray]{0.95} & Train all model weights together using CD on multi-frame samples of the data \cellcolor[gray]{0.95}\\ 
\end{tabular}
\end{center}
\end{table}

\subsubsection{Temporal Autoencoding for the CRBM}

The procedure is very similar for the CRBM. First the static weights $\mathbf{W}$ are initialized with contrastive divergence training. After that, we reconstruct the present frame from its past observations by passing it through the hidden layer. The obtained reconstruction is then a function of the past observations and the matrices $\mathcal{W}$ and the biases $\mathcal{B}$, we can write $\mathbf{\hat{v}}^T(\mathbf{v}^0_d,\mathbf{v}^1_d, \ldots , \mathbf{v}^{T-1}_d; \mathcal{W,B})$. We then perform stochastic gradient descent on the reconstrucion error
$$
\mathcal{L}(\mathcal{W,B}) = \frac{1}{Q}\sum_d \left\| \mathbf{v}_d^T -\mathbf{\hat{v}}^T(\mathbf{v}^0_d,\mathbf{v}^1_d, \ldots , \mathbf{v}^{T-1}_d; \mathcal{W,B})\right\|^2.
$$
After this step is finished we proceed to train the CRBM with normal contrastive divergence to fine tune the weights for better generation. A summary for the training procedure is given in \fref{tab:training} and a complete description of the temporal autoencoding step is given in \fref{alg:pretraining_crbm}.

\subsubsection{Implementation}

Gradient descent on the cost functions explained above involves backpropagation through the hidden layers. This has been made relatively simple by automatic differentiation packages such as Theano \citep{theano}. We have implemented the temporal autoencoding training as a MLP and then proceeded to perform stochastic gradient descent on the loss using mini-batches.

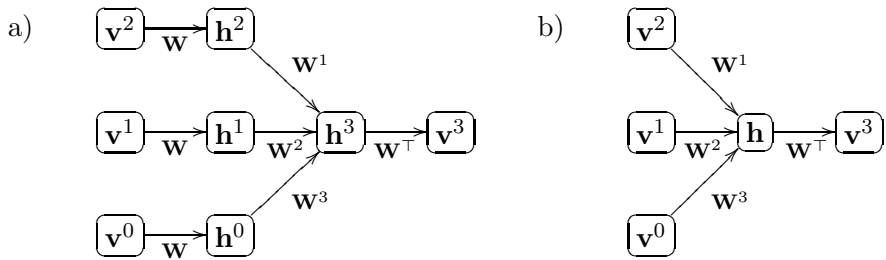
\begin{figure*}
\begin{displaymath}
\entrymodifiers = {+[F-:<3pt>]}
\xymatrix{
*{\textrm{a)}}& \mathbf{v}^2 \ar@{->}[r]_{\mathbf{W}}& \mathbf{h}^2\ar@{->}[dr]^{\mathbf{W}^1} &*{}&
*{}&*{\textrm{b)}}&\mathbf{v}^2 \ar@{->}[dr]^{\mathbf{W}^1}&*{}&*{}
\\
*{}& \mathbf{v}^1 \ar@{->}[r]_{\mathbf{W}}& \mathbf{h}^1\ar@{->}[r]_{\mathbf{W}^2} &\mathbf{h}^3\ar@{->}[r]_{\mathbf{W}^\top}&\mathbf{v}^3&*{}&\mathbf{v}^1\ar@{->}[r]_{\mathbf{W}^2} & \mathbf{h} \ar@{->}[r]_{\mathbf{W}^\top}&\mathbf{v}^3
\\
*{}& \mathbf{v}^0 \ar@{->}[r]_{\mathbf{W}}& \mathbf{h}^0\ar@{->}[ur]_{\mathbf{W}^3} &*{}&*{}&*{}&\mathbf{v}^0 \ar@{->}[ur]_{\mathbf{W}^3}
}
\end{displaymath}

\caption{TRBM (a) and CRBM (b) temporal autoencoding architectures.}
\label{fig:autoencoding}
\end{figure*}

\begin{algorithm}[htb]
\caption{Pre-Training Temporal weights through Autoencoding for the TRBM\label{alg:pretraining}}
\begin{algorithmic}
\STATE{Given a learning rate $\eta$}
\FOR{each sequence of data frames $I(t-T),I(t-(T-1))\ldots,I(t)$}
\STATE{take $\mathbf{v}^T = I(t), \ldots, \mathbf{v}^0 = I(t-T)$ and}
\FOR{$j = 1$ \bfseries{ to } $T$}
\FOR{$i=1$ \bfseries{ to } $j$}
\STATE $h^{T-i}_l = \sigma(\sum_k W_{kl}\,v^{T-i}_k+ b^h_l)$
\ENDFOR
\STATE $h^T_l = \sigma(\sum_{j=1}^T \sum_m W^j_{lm} h^{T-j}_m+b^h_l)$
\STATE $\hat{v}^T_n = \sigma( \sum_l W_{nl}\, h^T_l + b^v_n)$\\
$\mathbf{\epsilon}(\mathbf{v}^T,\hat{\mathbf{v}}^T) = |\mathbf{v}^T-\hat{\mathbf{v}}^T|^2$\\
$\Delta \mathbf{W}^d  = \eta \, \partial \epsilon/\partial \mathbf{W}^d$
\ENDFOR
\ENDFOR
\end{algorithmic}

\end{algorithm}

\begin{algorithm}[htb]
\caption{Pre-Training Temporal weights through Autoencoding for the CRBM\label{alg:pretraining_crbm}}
\begin{algorithmic}
\STATE{Given a learning rate $\eta$}
\FOR{each sequence of data frames $I(t-T),I(t-(T-1))\ldots,I(t)$}
\STATE{take $\mathbf{v}^T = I(t), \ldots, \mathbf{v}^0 = I(t-T)$ and}
\FOR{$j = 1$ \bfseries{ to } $T$}
\STATE $h_l = \sigma(\sum_{t=T-j}^{T-1} \sum_k W^{(T-t)}_{kl}\,v^{t}_k+ b^h_l)$
\STATE 
$\hat{v}^T_i = \sigma( \sum_l W_{il} \,h_l + b^v_i)$\\
$\mathbf{\epsilon}(\mathbf{v}^T,\hat{\mathbf{v}}^T) = |\mathbf{v}^T-\hat{\mathbf{v}}^T|^2$\\
$\Delta \mathbf{W}^d  = \eta \, \partial \epsilon/\partial \mathbf{W}^d$
\ENDFOR
\ENDFOR
\end{algorithmic}

\end{algorithm}

\section{Experiments}

We have applied our pretraining method to the CRBM and TRBM using two datasets. The motion-capture data described in \citep{taylor2007modeling} and the M3 competition dataset \citep{makridakis2000m3}. For both datasets we separated the data into a training and a validation set, then trained our models on the training set and evaluated them on a filling-in-frames task on the validation set. For all experiments we used a Gaussian-binary RBM model with variance fixed to 1.

\subsection{Motion-Capture Data}

We assessed the impact of our pretraining method by applying it to the 49 dimensional human motion capture data described in \citep{taylor2007modeling} and using this as a benchmark, comparing the performance to the models without pretraining.\footnote{In this section we refer to the reduced TRBM model referenced in \citep{sutskever2008recurrent} with only hidden-to-hidden temporal connections} All the models were implemented using Theano \citep{theano}, have a temporal dependence of 6 frames and were trained using minibatches of 100 samples for 500 epochs.\footnote{For the TRBM and CRBM, training epochs were broken up into 100 static pretraining and 400 epochs for all the temporal weights together. For the TA pretrained models, aTRBM and aCRBM, training epochs were broken up into 100 static pretraining, 50 Autoencoding epochs per delay and 100 epochs for all the temporal weights together, totalling to the same number of training epochs (500)} The training time for the models was approximately equal. Training was performed on the first 2000 samples of the dataset after which the models were presented with 1000 snippets of the data not included in training set and required to generate the next frame in the sequence. The generation in the TRBM is done using the filtering approximation, that is, by taking a sample from the hidden layers at $t-6$ through $t-1$ and then Gibbs sampling from the RBM at time $t$ while keeping the others fixed as biases. The visible layer at time $t$ is initialized with noise and we sample for 100 Gibbs steps from the model. The results of a single trial prediction for 4 random dimensions of the dataset can be seen in Figure \ref{fig:awesome} and the mean squared error and standard deviations of the model predictions over 100 repetitions of the task can be seen in Table \ref{tbl:motion}.\par
The models trained with Temporal Autoencoding significantly outperform their CD-only trained counterparts. The CRBM shows an improvement of approximately 56\%, while the TRBM shows an improvement of almost 80\% on this dataset. The performance can be further improved by taking the mean of the estimate by sampling from the hidden layer multiple times and taking the average prediction. This is akin to taking the Bayesian posterior mean estimator and leads to a further decrease in the MSE of 78\% for the CRBM and 91\% for the TRBM relative to straight CD training.
\par
One could argue that the improved performance of the TA pretrained model simply shows that a deterministic neural network is more well suited to the task at hand. To make sure that the improvement in performance is due to the interplay of both training approaches, we have also trained a deterministic multi-layer perceptron (MLP) with the architecture shown in \fref{fig:autoencoding}. This is shown in the rightmost column in \fref{fig:awesome}. As is shown, this simple deterministic approach outperforms the CD-trained model, but not the model trained with Temporal Autoencoding.
\par
These improvements also hold for longer time scales if we keep feeding the models predictions back into the it and let it generate autonomously. The TA pretraining significantly lowers the prediction error. Even after 6 frames, when all the visible layer frames were generated by the model, the MSE is still approximately as low or lower than when filling in one frame from the data without pretraining. The prediction errors for our models are shown in \fref{fig:prediction}.

\begin{figure*}[htb] \begin{center}
\includegraphics[width=\columnwidth]{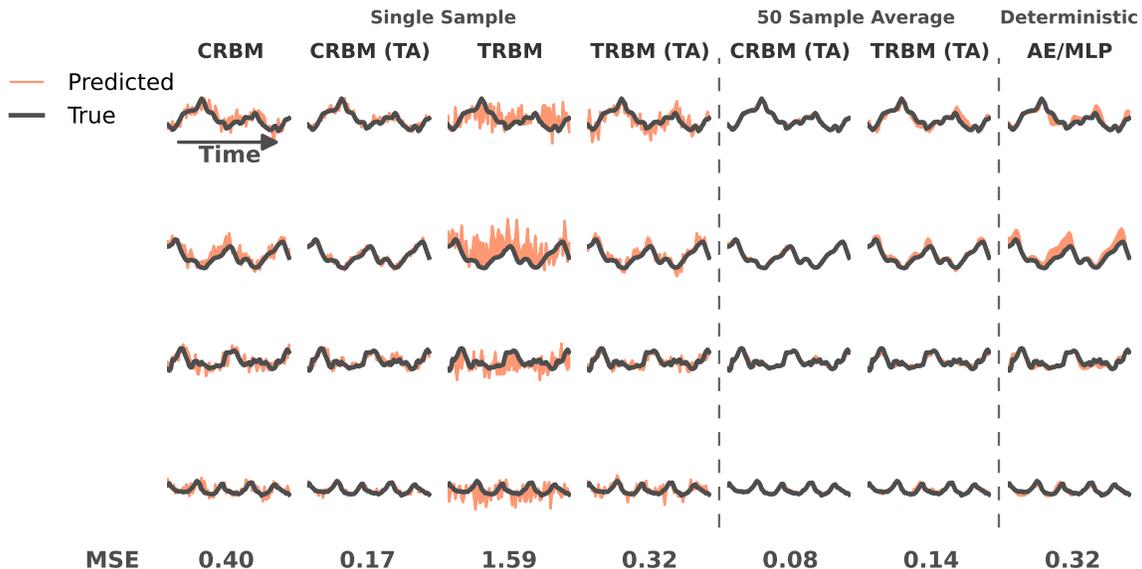}
\caption{The CRBM and TRBM are used to fill in data points from motion capture data \citep{taylor2007modeling} with and without TA pretraining. 4 dimensions of the motion data are shown along with the their model reconstructions from a single trial (left group), mean prediction over 50 samples (middle group) and deterministically (right group).} \label{fig:awesome} \end{center}
\end{figure*}

\begin{table}[t]
\caption{Prediction results on the motion capture dataset}
\label{tbl:motion}
\begin{center}
\begin{tabular}{lll}
{\bf Model} &{\bf Architecture and Training} &{\bf MSE ($\pm$ SD)}
\\ \hline \\
TRBM & 100 hidden units, 6 frame delay & 1.59 ($\pm\,0.12$)\\
TRBM (TA) & 100 hidden units, 6 frame delay & 0.32 ($\pm\,0.03$)\\
TRBM (TA), 50 sample mean & 100 hidden units, 6 frame delay & 0.14 ($\pm\,0.03$)\\
CRBM  & 100 hidden units, 6 frame delay & 0.40 ($\pm\,0.05$) \\
CRBM (TA) & 100 hidden units, 6 frame delay & 0.17 ($\pm\,0.02$)\\
CRBM (TA), 50 sample mean & 100 hidden units, 6 frame delay & 0.08 ($\pm\,0.02$)
\end{tabular}
\end{center}
\end{table}

\begin{figure*}[htb]\begin{center}
\includegraphics[width=.7\columnwidth]{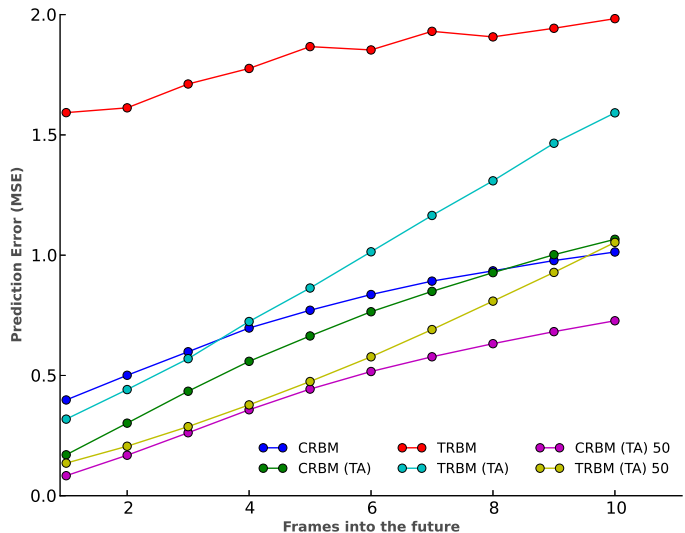}
\caption{CRBM, TRBM are used to fill in data points from motion capture data \citep{taylor2007modeling} with and without TA pretraining. The plot shows the evolution of the MSE after the input is killed and the model is left to generate samples on its own.} \label{fig:prediction} \end{center}
\end{figure*}

\subsection{M3 Forecasting Competition Data}

The motion capture experiments have shown great results for our proposed training method, but it reflects a lot of structure specific to the origin of the data. To assess how the method works on a more generalised dataset, we applied it to the datasets of the M3 forecasting competition.
The M3 forecasting competition \citep{makridakis2000m3} pitted forecasting algorithms against one another on 3003 different datasets, ranging from microeconomical to financial and industrial data. The data are univariate, but through state augmentation we can use our method to generate predictions for future data points. We have done so by taking chunks of 4 observations and used successive chunks as our multivariate data. With these we have trained the model to generate forecasts.\par
\Fref{fig:m3_prediction} shows the average performance of our algorithm on the four different kinds of data. They are separated into yearly, quarterly, monthly and other, the main categories of the competition. Here we measure the model performance using MAPE as was used in the competition. Although the datasets are generally small if compared to the usual unsupervised learning case, our training method still fares relatively well. Furthermore, TA pretraining continues to show a strong improvement over straight CD learning across the board. The robust performance of the TA pretraining on these datasets strongly suggests our method will generally yield improvements.

\begin{figure}[htb] \begin{center}
\includegraphics[width=.7\columnwidth]{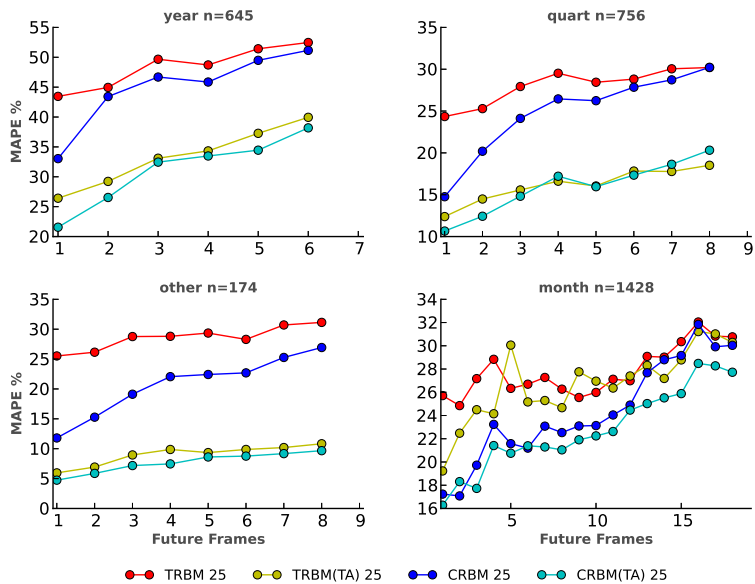}
\caption{CRBM, TRBM are used to fill in data points from the M3 forecasting competition with and without TA pretraining. The plot shows the evolution of the MAPE after the input is killed and the model is left to generate samples on its own. In all four data categories, the Temporal Autoencoded models out perform those without TA training} \label{fig:m3_prediction} \end{center}
\end{figure}

\section{Discussion and Future Work}

We have introduced a new training method for temporal RBMs we call Temporal Autoencoding and have shown that it can achieve a significant performance increase in a filling-in-frames task across a number of datasets. The gain in performance from our pretraining holds for both the CRBM and the TRBM, allowing for more efficient training of generative models.
\par
Our approach combines the supervised approach of backpropagating prediction errors through the network with the unsupervised approach of Contrastive Divergence learning. We have also shown that neither method by itself can achieve the performance we achieve by combining both.
\par
The approach shows significant improvement in the performance of the generative models, for filling-in-frames as well as for prediction tasks. This is shown to hold across a number of datasets. In the M3 contest dataset, specifically, the approach is shown to consistently improve the MAPE in a forecasting task, across a number of different types of data. On motion capture data, on the other hand, we were able to improve the MSE of the generative model by as much as 90\% in some cases.
\par
It is our opinion that the approach of autoencoding the temporal dependencies gives the model a more meaningful temporal representation than is achievable through contrastive divergence training alone. The TA training seeks to constrain the model to reproduce the dynamics observed in the data, as such it is not surprising that the improvement in generation also leads to an improvement in the prediction performance of the models considered.
We believe the inclusion of Autoencoder training in temporal learning tasks will be beneficial in a number of contexts, as it enforces the causal structure of the data on the learned model.
 \par


\bibliography{library}

\end{document}